\documentclass{isprs} 
\usepackage{subfigure}
\usepackage{setspace}
\usepackage{geometry} 
\usepackage{epstopdf}
\usepackage[labelsep=period]{caption}  
\usepackage[british]{babel} 
\usepackage[hang]{footmisc}

\usepackage{graphicx} 

\usepackage{color}
\usepackage{colortbl}

\usepackage{svg}

\begin{document}

\title{SLAM for indoor mapping of wide area construction environments}

\date{}

\author{V. Ress \thanks{Corresponding author}\enspace, W. Zhang, D. Skuddis, N. Haala, U. Soergel}

\address{Institute for Photogrammetry and Geoinformatics, University of Stuttgart, Germany -  forename.lastname@ifp.uni-stuttgart.de}


\icwg{II/Ib Digital Construction: Reality Capture, Automated Inspection, and Integration to BIM}   

\abstract{
Simultaneous localization and mapping (SLAM), i.e., the reconstruction of the environment represented by a (3D) map and the concurrent pose estimation, has made astonishing progress. Meanwhile, large scale applications aiming at the data collection in complex environments like factory halls or construction sites are becoming feasible. However, in contrast to small scale scenarios with building interiors separated to single rooms, shop floors or construction areas require measures at larger distances in potentially texture less areas under difficult illumination. Pose estimation is further aggravated since no GNSS measures are available as it is usual for such indoor applications. In our work, we realize data collection in a large factory hall by a robot system equipped with four stereo cameras as well as a 3D laser scanner. We apply our state-of-the-art LiDAR and visual SLAM approaches and discuss the respective pros and cons of the different sensor types for trajectory estimation and dense map generation in such an environment. Additionally, dense and accurate depth maps are generated by 3D Gaussian splatting, which we plan to use in the context of our project aiming on the automatic construction and site monitoring.}

\keywords{SLAM, Dense Reconstruction, Mobile Robot, BIM, Construction Sites, Close Range Sensing}

\maketitle


\section{Introduction}\label{MANUSCRIPT}
 
\begin{figure*}[h!]
\begin{center}
    \includegraphics[width=\linewidth]{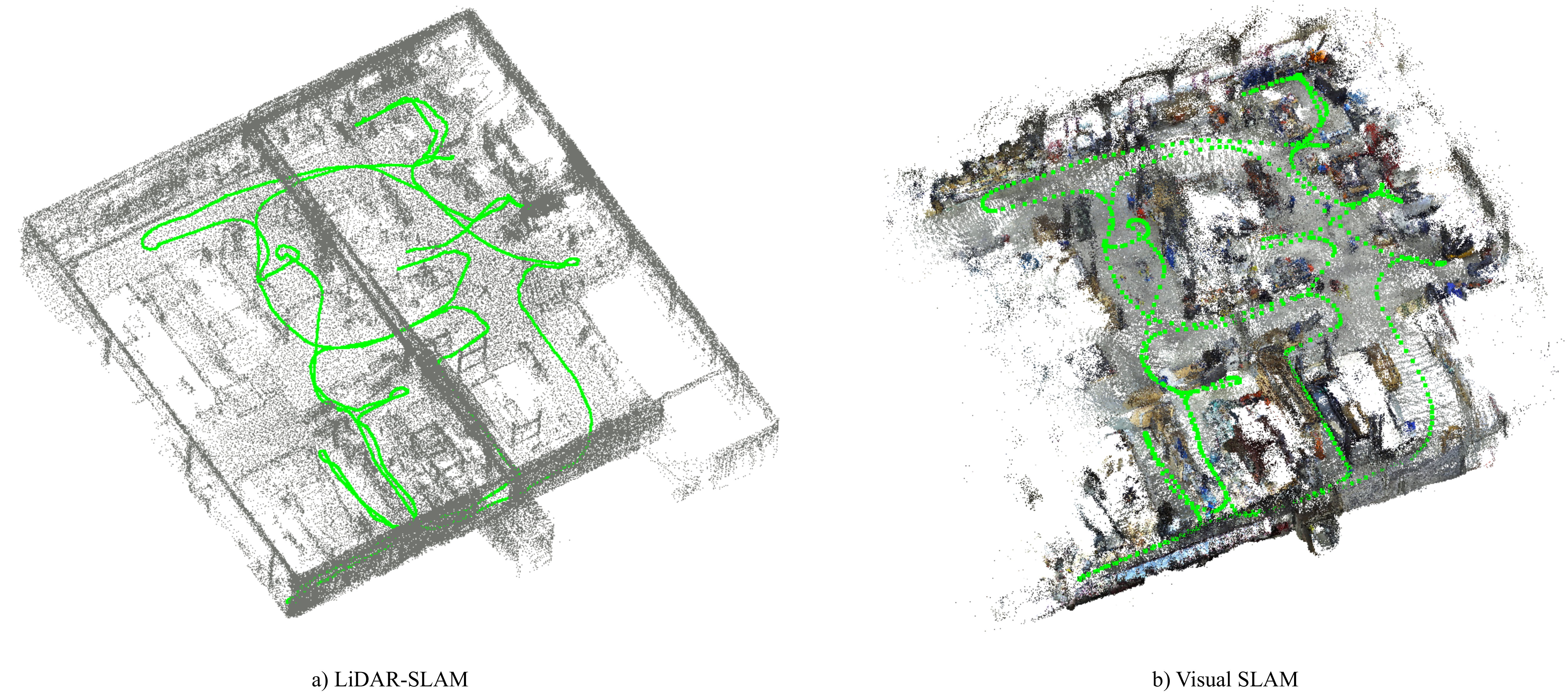}
	\caption{Visualisation of the resulting point clouds and trajectories for the LiDAR and Visual SLAM approaches presented.}
\label{fig:comparison_trajectory}
\end{center}
\end{figure*}

\sloppy
The problem of estimating the exterior orientation of optical sensors and the simultaneous reconstruction of the three-dimensional (3D) environment is commonly known as SfM (Structure from Motion) in Computer Vision and SLAM (Simultaneous Localisation and Mapping) in robotics \cite{Cadena2016}. In the context of SLAM, the estimation of the exterior orientation is performed sequentially and in real-time, frequently aiming at data collection in various indoor scenarios from images and/or laser scanning data.
The work presented in our paper is motivated by a project aiming at the monitoring of construction sites in the context of the cluster of Excellence Integrative Computational Design and Construction for Architecture (IntCDC) at the University of Stuttgart \cite{IntCDC}. Overarching goal of this work is to harness the potential of digital technologies to manufacturing and construction in the building sector. Construction industry has traditionally been a labor-intensive branch, yet it stands to benefit from autonomous robots that promise to deliver construction work that is more accurate and efficient compared to manual or conventional methods. One key task in this context is the direct digital data capture and monitoring of construction sites e.g. for the generation of BIM and digital twins. This documentation is an important application scenario for the geodetic capture of 3D point clouds. While in the past, Terrestrial Laser Scanning (TLS) from fixed stations defined the standard approach, meanwhile mobile systems applying SLAM-based methods are increasingly used. The state-of-the-art for data collection in such scenarios, is e.g. demonstrated by the results of the Hilti SLAM Challenge \cite{Hilti}.  For this benchmark, data acquisition was carried out by a hand-held system equipped with an IMU, a multi-camera head and a laser scanner device \cite{Zhang_Hilti}. According to the applied sensor type, SLAM algorithms are categorised into LiDAR and visual SLAM. Current visual SLAM methods can provide dense representations reasonably well, but are typically limited to well-textured environments and rather small spaces, such as single rooms, which are typically captured at short measurement distances. In contrast, real-world applications of large indoor scenes like construction sites and factory halls remain challenging and potentially benefit from the larger measurement range of LiDAR scans.  Figure \ref{fig:comparison_trajectory} exemplarily visualizes a reconstructed point cloud and trajectory for the LiDAR and Visual SLAM approaches as discussed in our paper. 

While approaches limited to a planar 2D space are quite mature and already enter consumer market, methods using 3D point clouds are still topic of research; however, such efforts are strongly pushed by applications connected with autonomous driving and robotics. Visual SLAM algorithms apply monocular, stereo or even RGB-D imagery. Compared to LiDAR sensors, cameras are significantly cheaper and therefore enable a much wider range of applications. Furthermore, the analysis of the captured images is not limited to geometric information extraction during localization and mapping. Due to rich information embedded in RGB imagery, visual SLAM is advantageous for visualization purposes and for semantic segmentation of the environment. On the other hand, the sensor principle of typical RGB-D devices limit their application to close range scenarios, while monocular SLAM frequently suffers from scale drift \cite{ORB2}. Even though the angular resolution of LiDAR sensors declines proportional to distance caused by diffraction, in practice, the direct range measurement principle allows for larger scene extent compared to visual SLAM schemes. As a result, current benchmarks (cf. \cite{Hilti}) show that LiDAR-based methods achieve significantly higher accuracy in trajectory reconstruction. However, this depends strongly on the scanning pattern and the line density as defined by the spatial resolution of the sensor. Further improvements in terms of localization reliability and accuracy for both groups of methods can be achieved by fusion with additional and complementing observations such as IMU or odometer data. In order to support data collection in larger scale environments like construction sites and factory halls, we use measures both from LiDAR and stereo cameras. As an example, we chose a factory hall, in which we monitor indoor construction activities \cite{IntCDC_LCR}. An exemplary result of our results from LiDAR and visual SLAM is already depicted in Figure \ref{fig:comparison_trajectory}.

The remainder of the paper is organized as follows: the following section on related work first presents current benchmarks aiming at scenarios similar to our task: data collection at large scale complex and dynamic environments. Furthermore, LiDAR and visual SLAM approaches feasible for such applications are introduced. Section \ref{sec:methods} then presents data collection by or robot system (cf. Section \ref{sec:method:sensor_platform}) and our processing pipelines for LiDAR and visual SLAM (cf. Sections \ref{sec:methods:LiDAR_SLAM} and \ref{sec:methods:visual_SLAM}). Our results for data collected in a wide-area factory hall are then presented in section \ref{sec:results}. Since our future work aims on the combination of LiDAR and visual SLAM, we especially discuss the pros and cons of the respective sensors for data collection in such an environment.

\section{Related Work}\label{sec:Rel_Work}
The following chapter is divided into three sections. The first introduces typical data sets used for SLAM applications and highlights the most important properties. In this context, only public available sources are considered. Subsequently, both established and the latest approaches in the fields of LiDAR and Visual SLAM will be presented. 

\subsection{Large-Scale SLAM Benchmarks}\label{sec:Rel_Work:Datasets}
Within the last years multiple data sets targeting various applications in the area of computer vision, simultaneous localization and mapping and robotics have been created. The majority of resources predominantly focus on capturing compact regions such as small rooms and address either visual or LiDAR SLAM by providing either camera or LiDAR data. 
One of the most common large-scale multi-modal data collections used to evaluate corresponding SLAM approaches is the KITTI data set \cite{Geiger_KITTI}. It contains stereo images, point clouds acquired by a 360 degree LiDAR and readings from an Inertial Measurement Unit (IMU) attached to a car. Various benchmarks for the calculation of depth maps, odemetry estimations or object detection and tracking tasks have been derived based on this resource. A comparable and even larger collection of driving data including radar information and additional cameras for a 360 degree view is the nuScenes \cite{Caesar_nuscenes} set 
Both resources are providing high quality data from various sensors, but are limited to outdoor scenes captured solely on streets. 
The data collection, which is in parts most closely aligned with our use case, is the HILTI benchmark  \cite{Zhang_Hilti}. For three scenes of this  collection the data were captured with an mobile robot equipped with a 3D front LiDAR, four RGB-D cameras for a 360 degree surround view and an IMU. Since all scenes were taken on construction sites, even the environmental conditions are comparable. Nevertheless, the focus on the HILTI benchmark is laid on the accurate localization of the robots. Therefore, the trajectories driven are not optimized for the purpose of a detailed reconstruction of the building or the objects around. Furthermore, the robot only moved in an underground car park with controlled lighting conditions, so that there were hardly any glare or reflection effects. In addition, no work was carried out in the robot's field of vision,i.e., apart from the rover and an instructor following it the scene was static.


In the following sections, algorithms for visual and LiDAR SLAM applications connected with our work are presented and described in more detail.

\subsection{LiDAR SLAM} \label{sec:Rel_Work:SLAM:Lidar}
Although the field of LiDAR SLAM originates from 2D LiDAR sensors, we will focus on more recent 3D LiDAR SLAM approaches in this overview.
An early popular real-time approach of such kind is LOAM \cite{zhang2014loam}. Here, from the LiDAR point set planar and edge features are derived. The poses of the sensor are determined from a high-frequency odometry function and a low-frequency mapping function by minimizing the errors of corresponding features.
Newer approaches also include mechanisms such as Scan Context \cite{kim2018scan} or LoGG3D-Net \cite{vidanapathirana2022logg3d} to detect the revisiting of already mapped locations (loop closures) and to improve consistency through pose graph optimization \cite{behley2018efficient}, \cite{ramezani2022wildcat}.
In some modern methods, the points are still reduced to edge and planar features during preprocessing \cite{shan2020lio}, while in others the points are combined into disk-like surface elements \cite{behley2018efficient}, \cite{ramezani2022wildcat} or in so-called dense approaches simply (downsampled) points are used \cite{dellenbach2022ct}, \cite{xu2022fast}, \cite{vizzo2023kiss}, \cite{ICRA_David}. In modern LiDAR SLAM systems IMU data are used to increase robustness in situations with fast movements and to reduce the orientational drift through gravity estimations \cite{ramezani2022wildcat}, \cite{shan2020lio}, \cite{xu2022fast}, \cite{ICRA_David}.
While in IMU-free LIDAR SLAM approaches it is common to model the trajectory within one scan as linear motion \cite{dellenbach2022ct}, \cite{neuhaus2019mc2slam}, \cite{vizzo2023kiss}, IMU supported LiDAR inertial SLAM approaches enable higher resolution trajectory representations, which can result in improved accuracy \cite{ramezani2022wildcat}, \cite{shan2020lio}, \cite{xu2022fast}, \cite{ICRA_David}.
In recent works, for LiDAR SLAM Neural Network representations of the environment  have been proposed to obtain more consistent maps \cite{wiesmann2023locndf}, \cite{zhong2023shine}, \cite{deng2023nerf}. In benchmark datasets, however, they cannot yet achieve the accuracy of conventional methods \cite{Geiger_KITTI}, \cite{Zhang_Hilti}.

\subsection{Visual SLAM} 
\label{sec:Rel_Work:SLAM:visual}
With ORB-SLAM \cite{ORB}, an efficient, robust and real-time capable SLAM approach that includes both place recognition capabilities required for loop-closing and global optimizations combined with the ability for lifelong mapping, which is especially required for the exploration of large environments, was introduced. ORB-SLAM and its successors, ORB-SLAM2 \cite{ORB2} and ORB-SLAM3 \cite{ORB3} are using hand-crafted feature extraction algorithms and optimization methods focusing on improved tracking accuracy. This results in low computing requirements during operation, but also causes relatively sparse point clouds and a low level of detail in the created maps. 
More recent approaches such as DROID-SLAM \cite{teed_droid} are integrating Neural Networks and train them on various scenes to further increase the robustness by reducing accumulating drift and the number of failures due to loss of feature track. A fully differentiable method design allows to combine and tune Neural Network Layers, for example, for dense pixel matching or to update the camera poses, with standard algorithms, for instance, to perform global optimizations. In addition, a clear separation in a frontend, which performs time critical tasks on the input stream of the images, and a backend, in which the computing-intensive processes are outsourced, still allows real-time capability. The study by  \cite{zhang2022towards} demonstrates the high effectiveness of DROID-SLAM in robotic applications with planar motion.

Most recent approaches are addressing the visualization and representation of the resulting maps. Traditional SLAM methods employ voxel grids, point clouds, or mesh representations as scene representations to construct dense mapping. However, these schemes face serious challenges in obtaining fine-grained dense maps. Methods such as HI-SLAM \cite{Hi_SLAM} extend existing concepts to include Neural Radiance Fields (NeRF) \cite{mildenhall_nerf} as 3D representation of the environment. For this purpose, the estimates such as pose and depth values of the keyframes are used to incrementally optimize the corresponding weights of the integrated Neural Network. In addition, by Multiresolution Hash Encoding \cite{Mueller_instant_nerf}  the required training times can be reduced significantly allowing a fast update of the resulting neural map. Recently, 3D Gaussian splats have been proposed as efficient rendering technique of Radiance Fields for high-quality and dense mapping with low memory consumption \cite{3DGaussianSplatting}. Beyond its efficiency for high resolution image rendering, Gaussian splatting holds an explicit geometry scene structure and appearance, benefiting from the exact modeling of scenes representation. This technology has been rapidly applied in several fields, and it seems to be also very promising for subsequent 3D modelling as one long-term goal of our project.

\section{SLAM-based mapping of a large factory hall} \label{sec:methods}
Within the following sub-chapters the sensor system used to capture the environment is introduced and a exploration of the characteristics of the recorded building is provided (cf. Chapter \ref{sec:method:sensor_platform}). The LiDAR and visual SLAM approaches (cf. Chapter \ref{sec:methods:LiDAR_SLAM}/\ref{sec:methods:visual_SLAM}) adapted and evaluated within this work are subsequently outlined .

\subsection{Sensor Platforms and Data Acquisition} \label{sec:method:sensor_platform}
As test environment a part of the Large-Scale Construction Robotics Laboratory (LCRL) of the Cluster of Excellence IntCDC  
was chosen \cite{IntCDC_LCR}. 
The building consists of a large construction hall including multiple robotic building part prefabrication plants, instructor work spaces, material depots and traditional fabrication tools. The setup includes wide areas with open space as well as small corridors. Multiple structures built from various materials such as concrete, steel or wood are located within the recording area. To support the evaluation of the resulting point clouds, markers have been placed at different locations within the hall. The data acquisition took place during normal operation so on-site staff was passing the sensors and robots and objects moved during our recording.

\begin{figure}[h!]
\begin{center}
    \includegraphics[width=0.95\columnwidth]{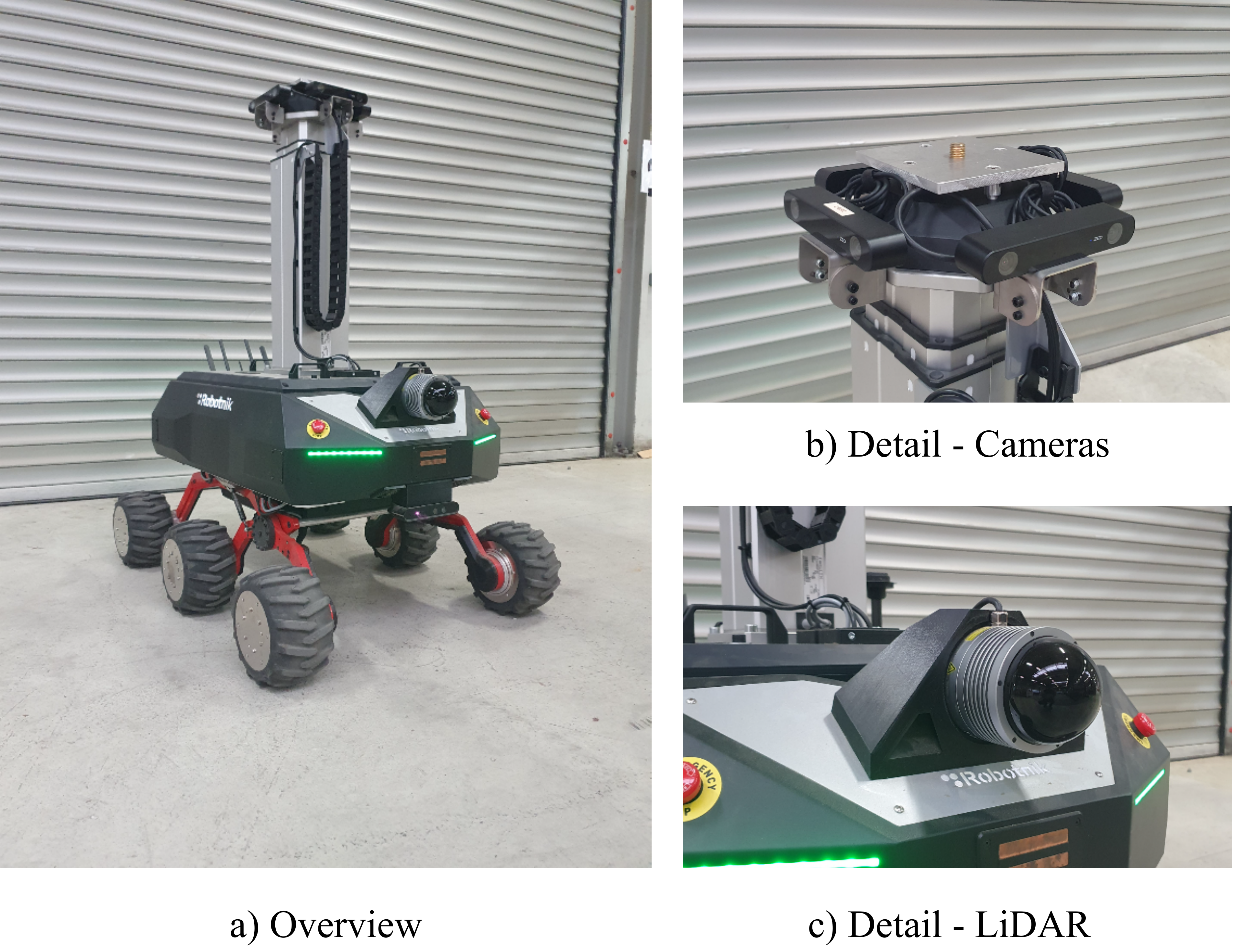}
    \caption{Images of the robotik platform used for data acquisition. }
\label{fig:robot}
\end{center}
\end{figure}

The sensor platform is a 6-wheeled robotic system providing all the basic functionalities such as power supply, computational resources and driving capabilities (cf. Figure \ref{fig:robot}). A 3D LiDAR sensor (RoboSense BPearl) with 32 lines and a maximum detection range of 100m (30m@10\% NIST) is mounted at the front of the robotic system. In addition, four  ZED 2 stereo cameras providing a 360$^{\circ}$-surround view of the environment are attached to an extractable tower in the center of the robot. All relative sensor orientations are known from prior calibration. Due to logistical reasons, only one third of the hall with the size of approx. $1600 m^2$ was explored by the robotic platform. For a most detailed reconstruction of the construction site, all areas that could be reached with the outer dimensions of the robot were entered. To provide a very accurate reference point cloud a Trimble X7 survey grade Terrestrial LiDAR Station was used. During data collection 360$^{\circ}$ scans were captured at eight scanning positions in the hall of the LCRL and merged to a final map using the available reference points.

\subsection{DMSA LiDAR SLAM}
\label{sec:methods:LiDAR_SLAM}

Within the scope of this work, we investigated three LiDAR SLAM methods to processes our recorded LiDAR / IMU data. In detail, we selected KISS-ICP \cite{vizzo2023kiss}, a popular LiDAR SLAM method that claims to be easy to integrate and robust, CT-ICP \cite{dellenbach2022ct}, a highly accurate LiDAR-only open source algorithm, and our DMSA SLAM \cite{ICRA_David}. While using the standard parameter settings proposed by the corresponding authors, the SLAM methods KISS-ICP and CT-ICP diverged after only a few seconds during processing. We suspect that the methods have difficulties with processing the very sparse LiDAR data of the RoboSense BPearl. Thus, the results presented in more detail in chapter \ref{sec:results} are solely generated by our DMSA SLAM approach. 

\begin{figure}[h!]
\begin{center}
		\includegraphics[width=0.90\columnwidth]{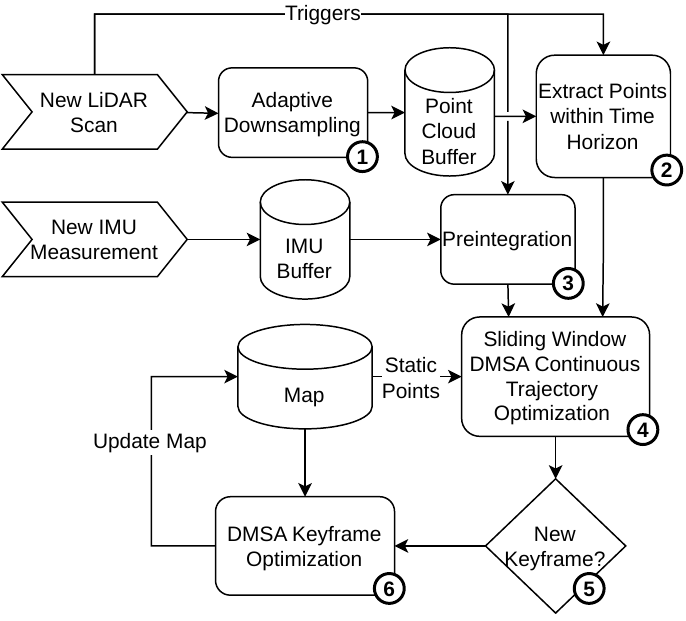}
	\caption{DMSA SLAM overview. Numbers indicate the processing order. Adopted from \protect\cite{ICRA_David}.}
\label{fig:dmsa_lio_float}
\end{center}
\end{figure}

In DMSA SLAM, LiDAR points within a sliding time window are optimized together with so called static points and IMU data in a tightly coupled manner. For processing the captured data within this work a sliding window time horizon of 0.8\,s is selected. Adaptive downsampling within the preprocessing enables handling of LiDAR data from narrow spaces as well as from spacious places. New keyframes are selected based on overlap and distance thresholds. In addition to LiDAR points belonging to a keyframe, the gravity direction is estimated and added to the keyframe data. When a new keyframe is added to the map, all keyframes with significant overlap to the current keyframe are optimized. Figure \ref{fig:dmsa_lio_float} gives an overview of the processing steps. For details we refer to \cite{ICRA_David}.

\subsection{Dense Multi-Camera RGB-D SLAM Pipeline}\label{sec:methods:visual_SLAM}

\begin{figure}[h!]
\begin{center}
    \includegraphics[width=0.95\columnwidth]{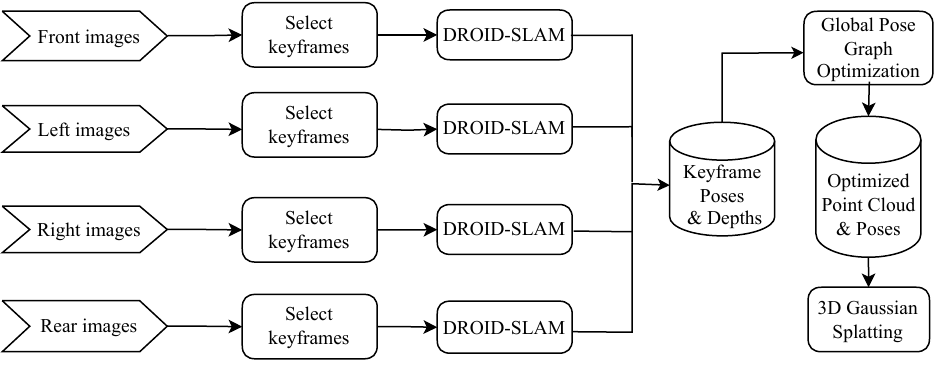}
    \caption{Diagram of dense visual SLAM pipeline.}
\label{fig:visual_pipeline}
\end{center}
\end{figure}

In the processing chain of the four camera imagery, we developed a dense visual SLAM pipeline visualized in Figure \ref{fig:visual_pipeline}. To fully leverage the 360-degree surround view, the data of each camera is first processed via the DROID-SLAM method \cite{teed_droid}. In this stage, keyframes for each camera view are adaptively selected based on the optical flow distance between neighboring images. For these keyframes, dense per-pixel depths are estimated by predicting dense flow and performing bundle adjustment per camera stream to minimize re-projection errors with the predicted flow as references. Subsequently, using the known extrinsics between the cameras, we transform the keyframe poses of each camera into a common coordinate system, with the optical center of the front camera as reference coordinate. 

Although this process yields a global map integrating all camera views and observations, inconsistencies may still arise due to remaining errors, such as drifted trajectories, in the parameters estimated individually for each camera - particularly in scenarios where no loop closure is available. To construct a globally consistent map, we combine the keyframes from all four cameras in a joint bundle block adjustment. For this purpose we adopted the global bundle adjustment of the DROID-SLAM backend. This optimization technique identifies loop closures across different camera views and observations at different times. 
The typical visual SLAM pipelines \cite{ORB3,teed_droid} often result in a point cloud map. However, this format may not be sufficient for robots to localize themselves within the map while navigating collision-free, due to the inherent sparseness of point clouds. Additionally, applications such as first-person navigation through a scene along a newly rendered path require realistic images from novel viewpoints. To overcome these challenges, we have adopted the 3D Gaussian Splatting method \cite{3DGaussianSplatting} for training a radiance field of the scene. This approach utilizes both the estimated poses and the globally consistent point cloud produced in the last stage as initial Gaussian positions. While following the default configuration in \cite{3DGaussianSplatting}, we have made a key enhancement by adding stereo depth for additional supervision, thereby improving geometric reconstruction. Figure \ref{fig:render_comp} presents an example of input stereo depth, which is often incomplete due to issues like low texture or limited stereo baseline. The depth images rendered using our trained 3D Gaussian model address these difficulties effectively, producing fully complete depths with clearer object borders.

\section{Results}\label{sec:results}
\begin{figure}[h!]
 \begin{center}
     \includegraphics[width=0.96\columnwidth]{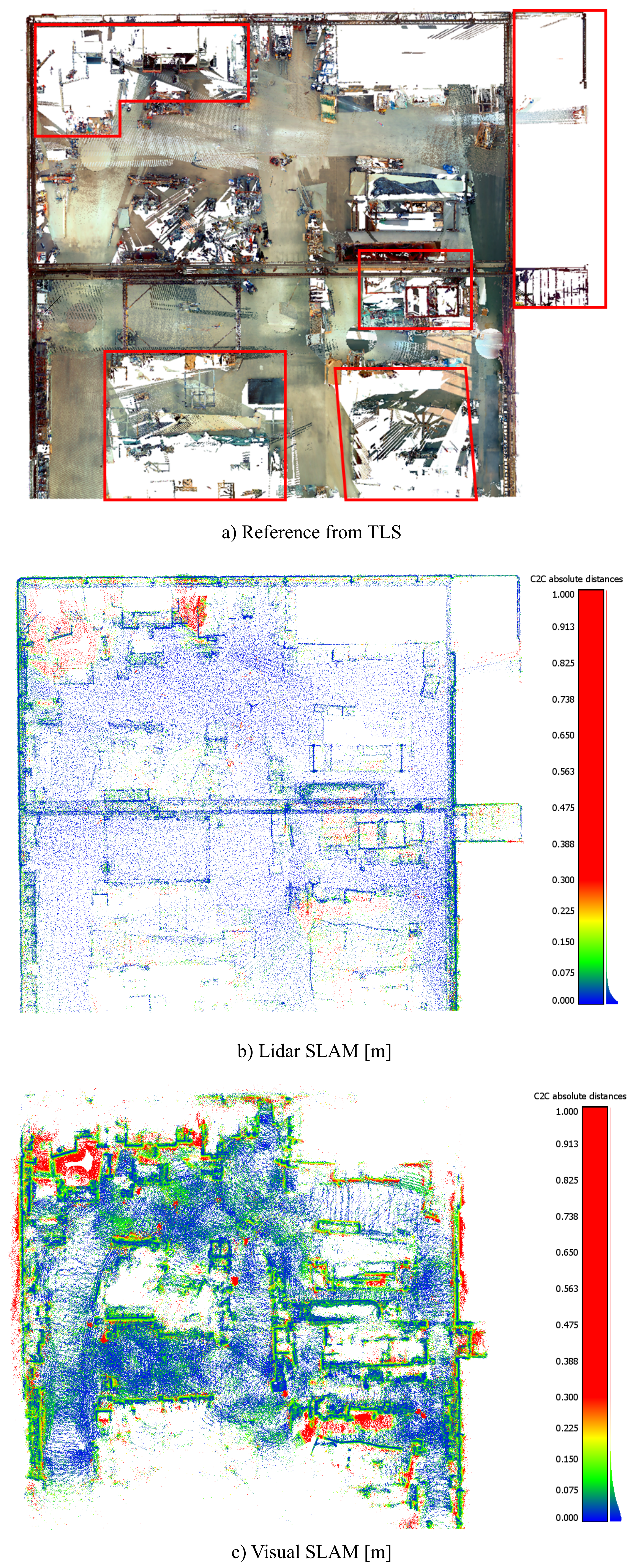}
     \caption{Visualization of the Cloud-to-Cloud (C2C) distance [m] of the presented DMSA LiDAR and RGB-D SLAM pipelines. Colorization based on the minimum distance of the points to the nearest neighbour of the reference point cloud (RGB).}
 \label{fig:comparison_map}
 \end{center}
 \end{figure}

In the following, we discuss the performance of the aforementioned methods step-by-step. While Figure \ref{fig:comparison_trajectory}a) and b) already presented the resulting point clouds including the determined trajectories from both our LiDAR and visual SLAM methods. Figure \ref{fig:comparison_map} further evaluates and visualizes these results based on an comparison to a TLS reference  (cf. Figure \ref{fig:comparison_map}a). To compare the created maps with the TLS reference, the corresponding points of each cloud were transformed to the reference coordinate system and then finely adjusted using Iterative Closest Point (ICP). The colorization of Figure \ref{fig:comparison_map}b and \ref{fig:comparison_map}c is based on the Euclidean distance to the nearest neighbour of the reference point cloud.


 \begin{figure*}[h!]
\begin{center}
    \includegraphics[width=0.87\linewidth]{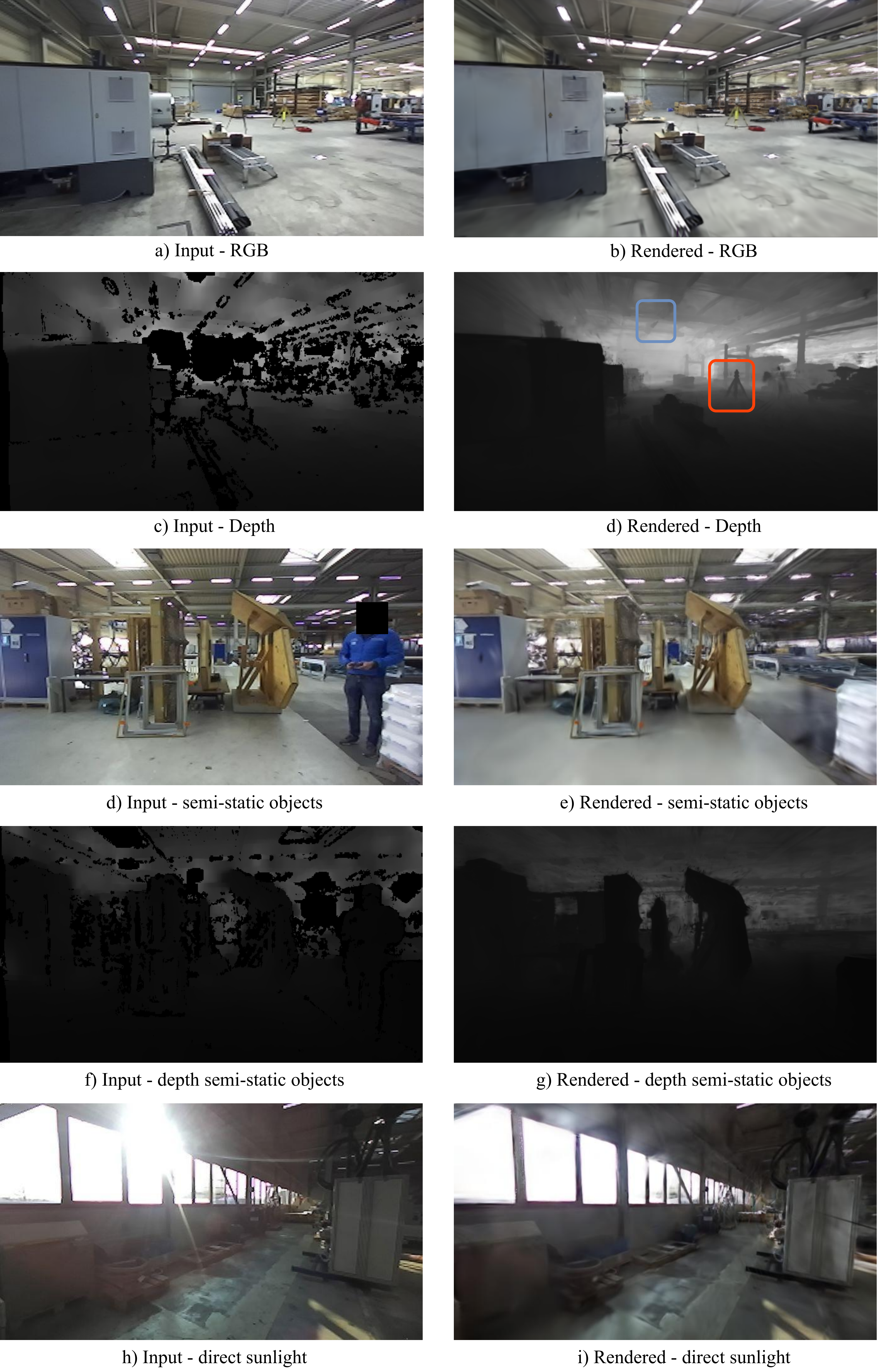}
	\caption{Comparison of input RGB and depth images with those rendered by our 3D Gaussian splatting model. Our reconstruction notably enhances depth quality and completeness. }
\label{fig:render_comp}
\end{center}
\end{figure*}

As the graphics illustrate, the point cloud resulting from the LiDAR SLAM approach (cf. Figure \ref{fig:comparison_map}b) provides a relatively sparse but precise representation of the environment. With DMSA SLAM \cite{ICRA_David} we were able to successfully process the data from the LiDAR sensor together with the IMU data after a few adaptations to the initially published pipeline/parameters. The alignment of the environment representation created on the basis of Visual SLAM showed that the resulting map was down-scaled by approximately 5\% compared to the reference. Scaling issues commonly arise in monocular SLAM, but can be addressed across various applications through the utilization RGB-D cameras \cite{ORB3}. However, since such cameras frequently apply stereo or structured light, they typically suffer from a restricted measurement range (max. $15-20m$) and, compared to LiDAR sensors, less precise depth information.  As an example, the applied ZED2 camera features a $120mm$ stereo baseline. According to the manufacturer, the maximum range of the ZED 2 is $20m$ before the depth’s accuracy decreases significantly. Our assumption is that particularly in wide area environments these limitations still contribute to (significant) scale errors. However, to simplify comparisons, for the following evaluations the scale of our visual SLAM result was corrected by a corresponding up-scaling of the affected point cloud. 

While LiDAR SLAM typically does not suffer from scaling issues and provides reliable range measures at considerable distances, our visual SLAM using all four RGB-D cameras  produces a denser point cloud after global optimization. However, it frequently suffers from a higher noise impact. This becomes particularly evident at the right and left walls of the hall, which look quite bumpy compared to the result achieved from LiDAR data. On the one hand, this is caused by the lower reliability of the depth estimates generated from the stereo camera, especially for distant objects, and on the other hand to challenging lighting conditions within the captured environment. 
Since both methods are not as much effected from shadowing effects as the reference point cloud measured from only a few dedicated scanning positions inside the hall, areas with no information in the reference cloud lead to high distances in the evaluation. These areas effect on the one hand the fine adjustment via ICP and on the other hand restrict possible quality assessments. For this reason we excluded those areas without a sufficient number of observations (cf. red boxes - Figure \ref{fig:comparison_map}a). Based on the remaining data we received a mean point distance of $\mu=4,1cm$ $(\sigma=6,8cm)$ for the point cloud created by the presented LiDAR SLAM approach and $\mu=7,4cm$ $(\sigma=9,5cm)$ for the resulting map of our visual SLAM pipeline.

Objects within the hall that are temporarily stationary (semi-static) and are thus located in the same place over several frames, result particularly in the point clouds of the camera-based method as local maxima (green/red dots) in the computed distances and thus appear as red or green clusters in the point cloud (c.f. Figure \ref{fig:comparison_map}b/c). Will these objects move from their first observed position, the corresponding points will remain as fragments in the generated point cloud. Examples are the operator following the robot or staff temporarily working at a certain location. Since the cameras of the robot are providing a $360^\circ$-view they are more sensitive to semi-static objects than the LiDAR sensor which is only scanning the front of the robot. 
In addition, the resolution of the LiDAR is significant lower than the data gathered by the camera. For these reasons,  semi-static objects in the point cloud generated by the BPearl sensor appear less frequent in comparison to the camera based approach.

The advantages of using 3D Gaussian Splats as representation of the environment is presented in Figure \ref{fig:render_comp}. By using Gaussian Splats a nearly photo-realistic representation of the environment can be created and even new perspectives, which have not been visited by the robot during data collection, can be rendered and visualized. Additionally, the combination of multiple observations lead to a significant improvement of the depth information. Thus, even small details such as the tripod of a tachymeter or the outlet of the air-condition system, which are not clearly visible on the depth map generated by the original sensor, become recognisable (cf. Figure \ref{fig:render_comp}c/d). The optimization process in the Gaussian splats approach may render certain semi-static objects invisible due to the occasionally redundant observation of the same building parts (cf. Figure \ref{fig:render_comp}d/e/f/g), which is also helpful for image rendering at poses suffering from challenging lighting conditions during image acquisition (cf. Figure \ref{fig:render_comp}h/i). The efficient rendering methods presented by \cite{3DGaussianSplatting} also enable the dynamic loading and real-time visualization of large environment models on typical consumer GPUs. We were able to attain update rates exceeding 30 fps using a NVIDIA GeForce RTX 3090 Ti. Furthermore, the improved depth information for each (virtual) station is very helpful for subsequent applications aiming at Augmented Reality or high precision navigation.

\section{Conclusions and Future Work}
As our main contribution we demonstrated the feasibility of SLAM-based methods for the acquisition and mapping of large factory halls or construction sites. Based on the data from our multi-sensor platform, we confirmed our original assumption that LiDAR sensors provide high geometric accuracy during the reconstruction of large scale environments, while camera-based methods can generate useful details which can be excellently visualised using 3D Gaussian splats. Compared to classic geodetic methods such as TLS, a number of advantages of SLAM-based methods for the mapping of large building interiors can be identified. In general, the flexible data collection of mobile platforms helps to avoid occluded areas especially in complex environments, which is typically labor-intensive for stationary measurement. Even though our SLAM pipeline does not (yet) allow for real-time processing, it already is very time efficient. The analyzed sequence of our factory hall depicted in Figure \ref{fig:comparison_trajectory} was captured in 0.3h, while TLS data acquisition took 1h for preparation and to 2h for acquisition. 
In simple terms, the more complex and finer the geometry of the environment, the greater the time savings for a mobile approach. In combination with a map-based trajectory planning and an automatic guidance even a fully autonomous acquisition process becomes imaginable.

While the achieved (mean) accuracies of our approach $(4-8cm)$ are sufficient for navigating through the captured environment, plenty of tasks on construction sites such as the installation of doors, windows or similar built-in parts require more precise measurements. Our future work thus aims on fusing LiDAR and image data while striving to get the best of both worlds - reliable position estimates and geometries from LiDAR SLAM in combination with rich representations of the environment from the visual SLAM. In order to obtain more reliable accuracy estimates for the resulting trajectories and maps, we also plan to integrate reference points located inside the hall into our evaluation process.

Since 3D Gaussian splats are characterized by efficient memory usage and, in combination with the already developed rendering methods, require low computing power, they provide an ideal opportunity of representing large environment models. Due to the explicit form of representation, they are also a good basis for subsequent 3D segmentation tasks \cite{garfield2024}. In our ongoing efforts to enhance the precision and quality of visualizations and 3D point clouds generated by our visual SLAM pipeline, we are also working on refining the joint calibration and photogrammetric assessment of the four cameras. If semantic information is also determined, digital construction plans such as Building Information Model (BIM) or Building and Habitats object Model (BoHM) can also be created or updated on the basis of the collected data. In this way, automatically generated building status reports that compare the actual status with the target status are conceivable.

\section{Acknowledgements}
Supported by the Deutsche Forschungsgemeinschaft (DFG, German Research Foundation) under Germany´s Excellence Strategy – EXC 2120/1 – 390831618.
{
	\begin{spacing}{1.17}
		\normalsize
		\bibliography{SLAM_ISPRS} 
	\end{spacing}
}

\end{document}